\documentclass[sigconf]{acmart}

\AtBeginDocument{%
  }

\usepackage{amsmath}
\usepackage{multirow}

\newcommand{\chan}[1]{\textsf{#1}}

\copyrightyear{2026}
\acmYear{2026}
\setcopyright{cc}
\setcctype{by}
\acmConference[PolDS '26]{The 2nd ACM SIGSPATIAL International Workshop on Polar Data Science}{November 03--06, 2026}{Riverside, CA, USA}
\acmBooktitle{The 2nd ACM SIGSPATIAL International Workshop on Polar Data Science (PolDS '26), November 03--06, 2026, Riverside, CA, USA}
\acmDOI{10.1145/3849740.3856203}
\acmISBN{979-8-4007-3062-7/2026/11}

\begin{document}

\title{Supraglacial Lake Fate Is Knowable Long Before the Season Ends}

\author{Emam Hossain}
\affiliation{%
  \institution{University of Maryland, Baltimore County}
  \city{Baltimore}
  \state{Maryland}
  \country{USA}}
\email{emamh1@umbc.edu}

\author{Md Osman Gani}
\affiliation{%
  \institution{University of Maryland, Baltimore County}
  \city{Baltimore}
  \state{Maryland}
  \country{USA}}
\email{mogani@umbc.edu}

\renewcommand{\shortauthors}{Hossain and Gani}

\begin{abstract}
A supraglacial lake on the Greenland Ice Sheet ends its melt season in one of
four ways: it drains rapidly through a hydrofracture, drains slowly across the
surface, refreezes in place, or is buried by late-season snowfall. Which one
occurs decides whether the meltwater reaches the ice bed. Satellite classifiers
recover the outcome accurately but only after the season closes, and how much of
a season each outcome actually requires has never been measured. We measure it
directly: holding the representation and the classifier fixed, we truncate the
input at $14$ cutoffs from May~1 to December~31, retrain at each, and record the
earliest cutoff at which each outcome's per-class $F_1$ reaches a fixed target.
The outcomes resolve in a consistent order, two of them months early: rapid
drainage by July~15 and slow drainage by August~1, respectively $92$ and $75$ days ahead of
the earliest date a full-season pipeline can be computed at all, with buried and
refreeze following at $44$ and $30$ days. Five further learners, from a
majority-class floor and $54$ summary statistics to a trigger-based early
classifier, leave the ordering largely intact: the three that produce a per-class
trajectory reproduce it in five of six cases despite end-of-season accuracies differing by up to
$18$ percentage points, and it survives leave-one-basin-out evaluation, though
not a move to machine-labeled lakes in an unseen season.
Every feature we compute at day $t$ reads
only days up to $t$, at a cost of at most
$1.3$ percentage points. A monitoring system should therefore not have one
release date: rapid drainage can be flagged on July~15, three months before a full-season pipeline can be computed at all.
\end{abstract}

\ccsdesc[500]{Computing methodologies~Machine learning approaches}
\ccsdesc[500]{Applied computing~Earth and atmospheric sciences}
\ccsdesc[300]{Information systems~Data mining}
\ccsdesc[300]{Mathematics of computing~Time series analysis}

\keywords{supraglacial lakes, Greenland Ice Sheet, early classification of time
series, temporal leakage, spatial cross-validation}

\maketitle

\section{Introduction}

\begin{figure*}[!h]
\includegraphics[width=\textwidth]{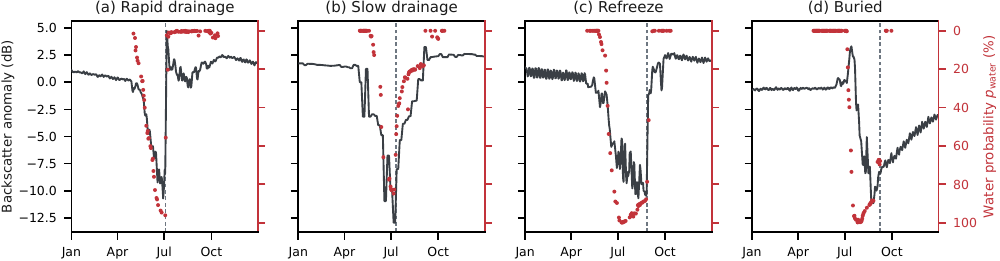}
\caption{One representative lake for each end-of-season outcome, from the raw
2019 record. Gray line: radar backscatter anomaly
\chan{HV\textsubscript{anom}} on the left axis, shared across panels. Red
points: merged water probability \chan{p\textsubscript{water}} on the right axis,
which runs downward so both channels move together when a lake loses its water.
The dashed line marks the last day the lake is observed as open water. The
backscatter trace is carried forward across gaps and smoothed over three days
for legibility.}
\Description{Four panels in a row, one per outcome class, each showing one
lake's 2019 record. Rapid drainage shows a sharp deep notch in backscatter with
immediate recovery. Slow drainage shows a broad decline and gradual recovery
over weeks. Refreeze shows an extended low through midsummer followed by a
step recovery at freeze-up. Buried shows a drop that persists with water still
detected into autumn.}
\label{fig:classes}
\end{figure*}

Surface meltwater ponds in thousands of supraglacial lakes across the Greenland
Ice Sheet each summer, and how each lake ends its season decides where that
water goes. A lake that drains rapidly, typically through a hydrofracture that
opens a moulin, delivers water to the ice bed within hours and can measurably
accelerate local ice flow~\cite{das2008fracture,zwally2002surface,
doyle2014persistent}. A lake that drains slowly routes its
water across the surface into an existing englacial system, without the same
dynamic signature~\cite{hoffman2018greenland}. A lake that refreezes holds its
water at the surface all winter, and a lake buried by late-season snowfall can
persist as a subsurface water body for
years~\cite{koenig2015wintertime,Forster2014ExtensiveSheet}. The four pathways
carry different consequences for ice dynamics and for sea-level
rise~\cite{shepherd2020mass,nienow2017recent,van2016recent}, and
Figure~\ref{fig:classes} shows how differently each is written into the record.

Satellite remote sensing has made these outcomes mappable at ice-sheet scale.
Dunmire et al.~\cite{dunmire2024greenland,dunmire2025agu} classify lake evolution across
Greenland by combining optical and synthetic aperture radar time series with a
manually labeled reference set, and other representations have been explored
for the task~\cite{hossain2024time,hossain2025rictsc}. All of these
systems share one design decision: every feature is computed over the whole melt
season, and the label is emitted once the season is over. For a retrospective
inventory that is the correct design, and the accuracies these systems report
are not in question here.

It leaves a different question unanswered. Retrospective labeling establishes
\emph{what} a lake did, not \emph{when} the observations settled it, and the gap
between the two is where the operational value lies. If a rapid drainage is
already unambiguous by mid-July, a system reporting it in October is late by a
margin belonging to the pipeline rather than to the ice. If refreeze genuinely
cannot be separated from a very late slow drainage until freeze-up, no modeling
will make an early refreeze label trustworthy. The two cases demand opposite
responses, and which holds for which outcome is unmeasured.

We ask that question under a deliberately conservative design. The
representation and classifier are fixed to a random convolutional transform and a
ridge classifier, which is fast, has one effective hyperparameter, and is among
the strongest general-purpose time series classifiers in published
benchmarks~\cite{dempster2021minirocket,middlehurst2024bakeoff}. We then vary
only how much of the season the model may see: truncating the input at $14$
cutoffs from May~1 to December~31 and retraining at each yields a per-class
accuracy trajectory, from which we read the earliest cutoff at which each
outcome reaches a fixed target. Because the answer should be a property of the
data rather than of a model, we repeat the sweep with five other learners, one
of them a majority-class floor and one purpose-built for early classification.
And because a lead time computed on features that read the future would not be
interpretable, we first rebuild
the preprocessing chain so that no feature at day $t$ depends on any observation
after day $t$, and we measure what that guarantee costs relative to the
conventional chain.

Our contributions are as follows.

\begin{itemize}
\item \textbf{Outcomes become knowable in order, and two of them early.} The
  four reach a fixed $F_1$ target in a fixed order, the two drainage outcomes
  reaching it $92$ and $75$ days before a full-season pipeline can be computed at
  all, so drainage and its absence should be released on different schedules
  (Section~\ref{sec:leadtime}).
\item \textbf{The ordering belongs to the data, not the model.} Six learners,
  from a majority-class floor and hand-built summary statistics to a
  trigger-based early classifier, run on identical folds; the four that produce a
  per-class trajectory recover the ordering in seven of eight learner-by-split
  cells despite end-of-season accuracies differing by up to $18$ percentage
  points, and it holds under leave-one-basin-out evaluation
  (Sections~\ref{sec:modelindep} and~\ref{sec:transfer}).
\item \textbf{Leakage-free preprocessing, and a measurement of what it costs.}
  Replacing gap interpolation, centered smoothing and season-wide statistics with
  strictly trailing equivalents makes the lead time interpretable for at most
  $1.3$ percentage points, so the correct way to compute one is also the cheap
  way (Section~\ref{sec:ablation-leak}).
\end{itemize}

\section{Background and Related Work}
\label{sec:related}

\subsection{Supraglacial lake evolution}
\label{sec:bg-lakes}

Supraglacial lakes form in topographic depressions on the ablation and lower
percolation zones of the ice sheet, filling from surface melt and from snowpack
drainage as the season advances~\cite{mcmillan2007seasonal,sundal2009evolution,
leeson2015supraglacial}. Their end-of-season fate is conventionally resolved into
four classes~\cite{dunmire2021contrasting,dunmire2024greenland}, illustrated in
Figure~\ref{fig:classes}.

\emph{Rapid drainage} is a hydrofracture event that empties a lake over hours to
a few days and opens a conduit to the bed. It is the outcome most directly
coupled to ice dynamics, because it delivers a pulse of water and establishes a
moulin that often persists into later seasons~\cite{das2008fracture,
chu2014hydrology,Catania2010PersistentSheet}. In the satellite record it appears
as a collapse of optical water fraction within days, with a sharp rise in radar
backscatter as the exposed lake bed roughens.

\emph{Slow drainage} empties a lake over weeks through supraglacial channels or
a pre-existing moulin~\cite{banwell2012modeling,dunmire2024greenland}. The same two
signals move, but gradually, so the rate rather than the event distinguishes it.

\emph{Refreeze} is the absence of drainage: the lake persists until air
temperatures fall and it freezes over~\cite{selmes2011fast,dunmire2024greenland}.

\emph{Buried} is the outcome when late-season snowfall covers a lake that has
not drained. The water can then persist beneath the surface for years, and buried
lakes are associated with firn aquifers and low-permeability ice slabs in the
percolation zone~\cite{koenig2015wintertime,Forster2014ExtensiveSheet,
schroder2020perennial,macferrin2019rapid,dunmire2022observations}.

Two properties of this taxonomy shape the rest of the paper. The classes carry
unequal operational value, since rapid drainage is the outcome an ice-dynamics
model most needs to know about and is also the one with the sharpest temporal
signature. And two of the four are defined partly by an event \emph{not}
occurring, which is inherently harder to establish early than an event that
does. This asymmetry is the physical reason to expect earliness to differ by
class, and it is why we measure it per class rather than in aggregate.

\subsection{Classifying outcomes from satellite data}
\label{sec:bg-ml}

Optical and radar records have been used to detect and delineate supraglacial
lakes for two decades~\cite{sundal2009evolution,selmes2011fast,
moussavi2015quantifying} and more recently to classify what becomes of them,
within a fast-growing literature on deep learning for remote sensing and Earth
system science~\cite{yu2024deep,lian2025recent,hossain2025correlation}. The systems closest to this work
differ in sensor, in how the temporal signal is summarized, and in what is
predicted.

One group estimates the state of a lake on a date and derives its outcome from
the completed state sequence. Williamson et al.~\cite{williamson2017supraglacial}
track area and volume from MODIS reflectance by thresholding and region growing,
declaring rapid drainage when a piecewise model fitted to the area curve drops
faster than a rate threshold. Benedek and Willis~\cite{benedek2021winter}
threshold seasonally composited Sentinel-1 scenes for the low backscatter of
liquid water beneath a frozen lid. Hochreuther et
al.~\cite{hochreuther2021fully} pair a normalized difference water index with a
random forest on Sentinel-2 and rebuild each lifecycle from the per-date masks,
and the same machinery has been carried to
Antarctica~\cite{moussavi2020antarctic}.

Dunmire et al.~\cite{dunmire2024greenland} produce the first ice-sheet-wide
classification of lake evolution into the four outcome classes used here,
combining Sentinel-1 HV backscatter with Sentinel-2 and Landsat-8 reflectance
over the 2018 and 2019 melt seasons, summarizing each lake's season into a fixed
feature vector, and fitting a stacked ensemble whose components are specialized
by sensor. Their manually labeled set of $1{,}000$ lakes is the ground truth
used in this paper. Two further studies change the representation. Hossain et
al.~\cite{hossain2024time} fit Gaussian mixture models to reconstructed phase
spaces of the Sentinel-1 and Sentinel-2 series, on a three-class version of the
problem, and report higher accuracy than LSTM-FCN~\cite{karim2018lstmfcn} and
other deep classifiers, and Hossain et al.~\cite{hossain2025rictsc} run joint
PCMCI+~\cite{runge2019detecting} over the channels to find per-basin causal
parents of the backscatter anomaly and restrict a MiniROCKET transform to them.

Every one of these systems consumes the season in full before emitting a label,
and none reports accuracy under truncated input. Table~\ref{tab:positioning}
places them on the axes that make the gap visible.

\begin{table}[!h]
\caption{Positioning. ``Temporal scope'' is the span of observations that a
feature available at prediction time may depend on. ``Truncated evaluation''
asks whether accuracy is reported as a function of how much of the season the
model has seen.}
\label{tab:positioning}
\footnotesize
\setlength{\tabcolsep}{4pt}
\begin{tabular}{@{}llll@{}}
\toprule
System & Sensors & Temporal scope & Truncated eval. \\
\midrule
Williamson et al.~\cite{williamson2017supraglacial}  & MODIS & full season & no \\
Benedek and Willis~\cite{benedek2021winter}          & S1    & full season & no \\
Hochreuther et al.~\cite{hochreuther2021fully}       & S2    & full season & no \\
Dunmire et al.~\cite{dunmire2024greenland}           & S1, S2, LS & full season & no \\
Hossain et al.~\cite{hossain2024time}                & S1, S2 & full season & no \\
Hossain et al.~\cite{hossain2025rictsc}              & S1, S2, LS, CARRA & full season & no \\
ELECTS~\cite{russwurm2023elects}                     & S2    & causal & one policy \\
\textbf{This work}                                   & S1, S2, LS, CARRA & causal & per class \\
\bottomrule
\end{tabular}
\end{table}

\subsection{Early classification of time series}
\label{sec:bg-ects}

Early classification of time series (ECTS) studies the trade-off this paper
measures: how little of a series suffices for a reliable label. Four families of
methods have emerged. \emph{Trigger-based} methods attach a stopping rule to a
conventional classifier: the ECTS algorithm sets a minimum prediction length per training
instance from nearest-neighbor stability~\cite{xing2012early}, ECDIRE learns
per-class timestamps after which predictions become
reliable~\cite{mori2017reliable}, and TEASER trains
a one-class classifier over the base classifier's output at each candidate
decision point and commits once a fixed number of consecutive points
agree~\cite{schafer2020teaser}. \emph{Cost-optimization} methods minimize a
weighted sum of misclassification and delay cost, by per-timestamp regression in
CALIMERA~\cite{bilski2023calimera} and by a continuous-time policy for
irregularly sampled series in Stop\&Hop~\cite{hartvigsen2022stophop}.
\emph{End-to-end} methods learn the stopping rule jointly with the classifier,
as in ELECTS~\cite{russwurm2023elects} and in the variational formulation of
Chen et al.~\cite{xie2020learning}. \emph{Calibrated} methods use conformal risk control
to bound the accuracy lost by stopping early~\cite{ringel2024early}, and a survey
documents the same tension across all four~\cite{gupta2020approaches}.

These methods answer a related but distinct question. Most optimize a
\emph{single} earliness policy for a dataset, one trigger or one cost trade-off
applied to every instance regardless of its class, whereas the question here is
how earliness \emph{differs across outcome classes}. We treat that as an
empirical claim rather than a definitional one and test it in
Section~\ref{sec:modelindep} by running TEASER on the same lakes and folds.

\subsection{Leakage in temporal pipelines}
\label{sec:bg-leak}

A lead-time claim is only as good as the guarantee that a feature available at
day $t$ depends on nothing after day $t$. Three preprocessing steps in common
use break that guarantee. Interpolation across gaps fills a missing observation
from the values that bracket it, the default in widely used imputation
toolkits~\cite{moritz2017imputets} and applied to lake records in several of the
systems above~\cite{hochreuther2021fully,dunmire2024greenland}. Centered
smoothing windows average across days on both sides of $t$, so a window
straddling a drainage event carries post-event information into pre-event
features. And season-wide statistics, computed over the complete record and
attached to every day of it, are the usual way to build a fixed-length feature
vector~\cite{dunmire2024greenland,williamson2017supraglacial}.

None of these is an error when the target is a retrospective inventory; they
become disqualifying only when the question turns temporal. This is an instance
of a broader failure mode in machine-learning-based
science~\cite{kapoor2023leakage} and of the difficulty of validating models on
temporally ordered data~\cite{bergmeir2012cv}, and it compounds with a second
problem: random cross-validation on spatially structured data places neighboring
instances on both sides of the split and inflates measured
accuracy~\cite{roberts2017crossvalidation}. We rebuild the preprocessing chain
so that every feature is a function of days $\le t$ (Section~\ref{sec:causal}),
treat leave-one-basin-out as the primary evaluation
(Section~\ref{sec:protocol}), and measure both choices rather than assume them
(Sections~\ref{sec:ablation-leak} and~\ref{sec:transfer}).

\section{Problem Setup}
\label{sec:setup}

\subsection{Data and labels}
\label{sec:data}

We use the Greenland Ice Sheet supraglacial lake record compiled by Dunmire et
al.~\cite{dunmire2024greenland} for the 2018 and 2019 melt seasons. That record
supplies the observations and the ground-truth outcome labels; the truncated
prediction protocol, the preprocessing chain, the splits and every result in
this paper are new. Lakes carry the drainage-basin assignment of that record,
which partitions the ice sheet into Central West (CW), Northeast (NE), North
(NO), Northwest (NW), Southeast (SE) and Southwest (SW); those six basins define
the spatial folds of Section~\ref{sec:protocol}.

Each lake $i$ is a multivariate daily series
$x_i \in \mathbb{R}^{C \times 365}$ on a common day-of-year grid, with the
$C = 9$ channels of Table~\ref{tab:channels}. Three carry the radar view.
Cross-polarized HV backscatter responds strongly to volume scattering in snow
and ice and weakly to smooth open water, so a water-covered lake appears dark
and an emptied, roughened bed appears bright, the contrast that underpins radar
lake and melt mapping on both ice sheets~\cite{nagler2015sentinel}. We take the
mean inside the lake polygon, the mean over a surrounding buffer, and their
difference \chan{HV\textsubscript{anom}}, which removes the regional seasonal
cycle. Three channels carry the optical view, the water-classified fraction of
the polygon in each optical sensor and a merged water probability. The last
three describe conditions rather than the lake itself: air temperature and the
two solar zenith angles, the latter encoding when an optical observation was
geometrically possible at all and so distinguishing a genuinely dry lake from an
unobservable one.

\begin{table}[!h]
\caption{The nine input channels, the input to every model reported in the
paper. Section~\ref{sec:ablation-carra} tests five further atmospheric channels
as an ablation and finds that they do not help.}
\label{tab:channels}
\small
\setlength{\tabcolsep}{4pt}
\begin{tabular}{@{}lll@{}}
\toprule
Symbol & Source & Quantity \\
\midrule
\chan{HV\textsubscript{lake}}   & Sentinel-1 & Mean HV backscatter within the lake polygon \\
\chan{HV\textsubscript{out}}    & Sentinel-1 & Mean HV backscatter in the surrounding buffer \\
\chan{HV\textsubscript{anom}}   & Sentinel-1 & Backscatter anomaly, lake minus buffer \\
\chan{S2\textsubscript{water}}  & Sentinel-2 & Water-classified fraction of the polygon \\
\chan{LS\textsubscript{water}}  & Landsat-8  & Water-classified fraction of the polygon \\
\chan{p\textsubscript{water}}   & S2 + LS    & Merged water probability \\
\chan{t2m}                      & Reanalysis & Near-surface air temperature \\
\chan{S2\textsubscript{zenith}} & Sentinel-2 & Solar zenith angle at overpass \\
\chan{LS\textsubscript{zenith}} & Landsat-8  & Solar zenith angle at overpass \\
\bottomrule
\end{tabular}
\end{table}

Ground-truth labels come from the manually labeled reference set released with
the record: $1{,}000$ lakes from the 2019 season, assigned by expert
interpretation of the paired optical and radar imagery, with exactly $250$ in
each class. That balance is a property of how the set was assembled rather than
of the ice sheet, and it fixes the trivial floor of the task at $25.0\%$
macro-recall. A much larger set of machine labels accompanies the
record; we use it only as a transfer target in Section~\ref{sec:transfer}, where
it supplies $5{,}146$ additional lakes in 2019 and $3{,}846$ in 2018. The
processed record begins at day-of-year $121$ (May~1), before which the optical
and atmospheric channels have zero variance across lakes, so day $121$ is a hard
floor on any cutoff that can be evaluated.

\subsection{Prediction at a truncated cutoff}
\label{sec:truncation}

Let $y_i \in \mathcal{Y}$ be the end-of-season outcome of lake $i$, where
$\mathcal{Y} = \{\textrm{rapid drainage}, \textrm{slow drainage},
\textrm{buried}, \textrm{refreeze}\}$. Define the truncation operator
\begin{equation}
  \Pi_T(x_i) \;=\; \big[\, x_i[:,1], \; x_i[:,2], \; \dots, \; x_i[:,T] \,\big],
\end{equation}
which retains the first $T$ days of the year and discards the remainder. For
each cutoff $T$ we learn a classifier
$\hat f_T : \mathbb{R}^{C \times T} \rightarrow \mathcal{Y}$ that sees truncated
inputs at training time and at test time alike. The cutoff grid is
\begin{equation}
  \begin{split}
  \mathcal{T} = \{&121, 135, 152, 166, 182, 196, 213,\\
                   &227, 244, 258, 274, 288, 304, 365\},
  \end{split}
\end{equation}
the 1st and 15th of each month from May~1 to October~15, plus October~31, which
closes the melt season, and day $365$, a full-year reference. Both study years
are non-leap years, so these day-of-year values map to identical calendar dates
in 2018 and 2019 and the cross-year comparison in Section~\ref{sec:transfer} is
exact.

We fit a separate model at each cutoff rather than evaluating one full-season
model on shortened inputs. A model trained on $365$-day inputs and tested on a
$152$-day input is penalized twice, once for the missing evidence and once
because the input no longer resembles its training distribution, and after the
fact the two cannot be separated. Only the first is of interest. Retraining
removes the second penalty and matches deployment, since an operator wanting a
label on July~15 would train on what was available by then.

\subsection{Knowability and lead time}
\label{sec:tstar}

Let $F_{1,c}(T)$ be the per-class $F_1$ of $\hat f_T$ on class
$c \in \mathcal{Y}$. The \emph{knowability date} of class $c$ at target $\tau$
is the earliest cutoff meeting that target:
\begin{equation}
  T^\ast_c(\tau) \;=\; \min\{\, T \in \mathcal{T} \;:\; F_{1,c}(T) \ge \tau \,\}.
\end{equation}
We report $\tau = 0.80$ throughout and sweep it in Section~\ref{sec:ablation-tau}.

\begin{figure}[!h]
\includegraphics[width=\columnwidth]{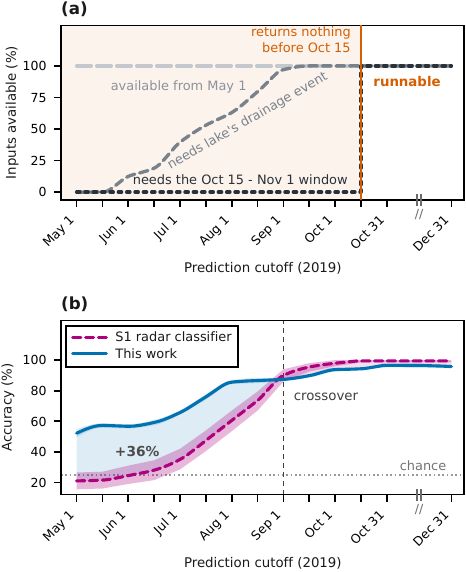}
\caption{(a) When each feature group of a full-season classifier becomes
computable. Nothing requiring October or full-season observations exists before
October~15, which fixes the reference date for lead time. (b) That classifier's
Sentinel-1 component on truncated input without retraining, against the model of
this paper retrained at each cutoff, both scored on the $200$ lakes the released
model holds out. The dashed vertical line marks where they cross.}
\Description{Two panels. The left panel shows availability curves for four
feature groups against date, with a vertical line at October 15 marking the
first date the full-season pipeline can be computed. The right panel shows two
accuracy curves against prediction cutoff, the released Sentinel-1 classifier
and the model of this paper, separated by about 36 points in mid-May, closing
through the season and crossing on September 1.}
\label{fig:incumbent}
\end{figure}

\begin{figure*}[!h]
\includegraphics[width=\textwidth]{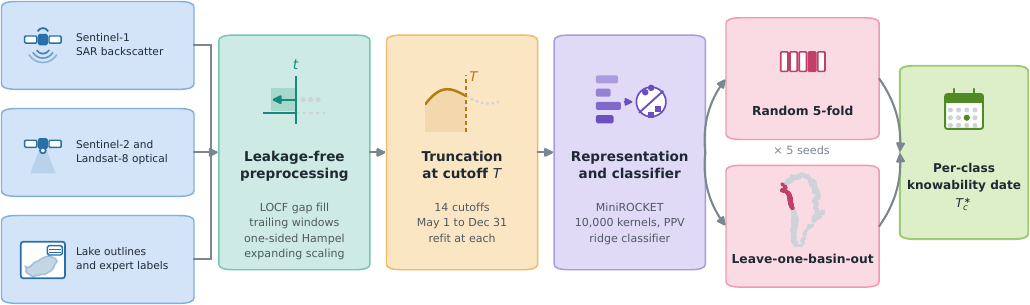}
\caption{The end-to-end architecture. Three input sources collect onto a common
daily grid and pass through the processing chain, which forks at the evaluation
protocol and converges on the quantity the pipeline emits. Each module type
carries its own color, and modules that do the same job share one: the three
input sources are a single family, as are the two split schemes. The
preprocessing glyph shows the rule the chain enforces, that a feature at day $t$
reads a trailing window and nothing after $t$. The leave-one-basin-out glyph is
not a schematic: it is the position of every lake in the 2019 record, projected
to polar stereographic and colored by drainage basin, with one basin held out.
}
\Description{A left-to-right wired diagram. Three input cards for Sentinel-1
radar, Sentinel-2 and Landsat optical, and lake outlines with expert labels
collect onto a vertical rail. The rail feeds three processing cards in
sequence: leakage-free preprocessing, truncation at a cutoff, and the
representation and classifier. These fork into two evaluation cards, random
5-fold and leave-one-basin-out, the latter drawn as a map of Greenland lake
positions with one basin highlighted. Both converge on a final card giving the
per-class knowability date.}
\label{fig:pipeline}
\end{figure*}

A knowability date is meaningful only against a reference date, and the
reference we adopt is a property of the input calendar rather than of any
particular system. A full-season classifier of the kind described in
Section~\ref{sec:bg-ml} draws on four groups of features: those available
continuously, those requiring the drainage event to have occurred, those
requiring October observations, and those requiring the complete record.
Figure~\ref{fig:incumbent}(a) traces when each becomes computable. The first is
usable from May~1 and the second fills in through the season, but the last two
are a step function that rises on October~15, so no classifier of this shape can
be evaluated at all before day-of-year $288$. The October window closes on
November~1, so this reference is conservative. We therefore define
\begin{equation}
  \mathrm{lead}_c(\tau) \;=\; 288 - T^\ast_c(\tau)
\end{equation}
in days: when information arrives relative to a retrospective label, not
relative to the event itself, and not a margin over a rival.

Figure~\ref{fig:incumbent}(b) shows the double penalty of
Section~\ref{sec:truncation} in that same classifier, on the $200$ lakes it
holds out. Without retraining, its Sentinel-1 component falls to $21.0\%$
accuracy on May~1 against a $25.0\%$ chance level, $31.2$ points below the model
of this paper retrained at the same cutoff; its Sentinel-2 component behaves the
same way (Appendix~\ref{app:results}). The gap closes through the season and
reverses on September~1: what the released model lacks early is evidence, not
capacity.

\section{Method}
\label{sec:method}

Figure~\ref{fig:pipeline} gives the end-to-end pipeline: the per-lake daily
record, the leakage-free preprocessing chain, the truncation operator, the fixed
classifier, the two split schemes and the reported quantity.

\subsection{Leakage-free preprocessing}
\label{sec:causal}

We rebuild the preprocessing chain so that every feature at day $t$ is a
function of observations on days $\le t$ only. Four substitutions do the work:
one for each of the three steps identified in Section~\ref{sec:bg-leak}, and one
for the outlier rejection that any moving statistic requires.

\textbf{Gap filling by last observation carried forward.} Cloud cover and orbit
geometry leave gaps of several days in the optical channels, and interpolation
fills a gap from the observations on both sides of it~\cite{moritz2017imputets}.
We instead carry the last observed value forward, the standard causal
alternative in online monitoring: the value at day $t$ is the most recent
measurement not later than $t$. Before a lake's first observation there is
nothing to carry forward and backward filling would again read the future, so
that stretch takes fixed constants from physical priors rather than from data
(Appendix~\ref{app:data}). Both chains use the same constants, so
Section~\ref{sec:ablation-leak} measures leakage, not fill.

\textbf{Trailing rather than centered smoothing.} A centered moving average of
width $w$ at day $t$ spans $[t - w/2, t + w/2]$, so with the $12$-day window
used here a feature two days before a drainage event already contains several
post-event days. The trailing average over $[t - w + 1, t]$ has the same
bandwidth and noise suppression but reads only the past. Its cost is a phase
lag, since a trailing average responds to a step change $w/2$ days later than a
centered one, which makes the measured dates conservative rather than optimistic.

\textbf{One-sided robust outlier rejection.} Speckle in the radar channels and
misclassified cloud edges in the optical channels produce isolated spikes that
distort any moving statistic. The Hampel filter replaces any sample more than $k$
scaled median absolute deviations from the local median, using the median as a
high-breakdown location estimate~\cite{hampel1974influence,pearson2016hampel}.
Its usual centered window would reintroduce the dependence just removed, so both
statistics use the trailing window.

\textbf{Expanding-window standardization.} Channel scales differ by orders of
magnitude, and season-wide statistics rescale every day using information from
every other day. We instead standardize day $t$ by the mean and standard
deviation over days $1$ to $t$.

Each substitution has a cost, and Section~\ref{sec:ablation-leak} measures
their total.

\subsection{Representation and classifier}
\label{sec:model}

Every truncated input $\Pi_T(x_i)$ is passed through
MiniROCKET~\cite{dempster2021minirocket} and classified by a ridge classifier.

MiniROCKET is a random convolutional kernel transform. It convolves the input
with a large bank of short kernels and summarizes each convolution by the
proportion of positive values (PPV), giving one feature per kernel. Unlike its
predecessor ROCKET~\cite{dempster2020rocket}, which samples kernel lengths,
weights, biases and dilations at random, it fixes almost all of these: kernels
have length $9$ with weights from $\{-1, 2\}$ in one of $84$ fixed patterns,
dilations follow from the series length, and only the biases are sampled from
the data. It is therefore nearly deterministic and roughly an order of magnitude
faster, while matching or exceeding ROCKET's accuracy on standard
benchmarks~\cite{dempster2021minirocket,middlehurst2024bakeoff}. We use
$10{,}000$ kernels over the nine channels jointly, giving $9{,}996$ features; the
transform is unsupervised.

The classifier is a ridge regression on one-hot targets, predicting the class
with the largest fitted score. Its closed-form solution suits the very wide,
low-sample regime the transform produces ($10^4$ features against at most $10^3$
training lakes), and its leave-one-out cross-validation error follows exactly
from a single singular value decomposition of the design
matrix~\cite{hastie2009elements}. We use that identity to select the
regularization strength within each training fold from ten logarithmically
spaced values in $[10^{-3}, 10^{3}]$, so no hyperparameter sees test data.
Implementations are sktime~\cite{loning2019sktime} and
scikit-learn~\cite{pedregosa2011scikit}.

Holding this pairing fixed across cutoffs is central to the design, since
changing the model while changing the amount of input would confound the two
effects. Section~\ref{sec:modelindep} then tests whether the choice affects the
conclusion, by repeating the entire sweep with the alternatives of
Section~\ref{sec:baselines}.

\subsection{Baselines}
\label{sec:baselines}

We compare against five further learners, each chosen to rule out a specific
alternative explanation of the measured result rather than to populate a
leaderboard. All run on identical folds, seeds, cutoffs, labels and preprocessed
inputs, so nothing but the learner changes.

\textbf{T1, majority class}, predicts the most frequent class in the training
fold. A constant prediction on four balanced classes attains $25.0\%$
macro-recall by construction; because the majority class differs from fold to
fold, T1 reaches $22.1\%$ under random folds and $16.6\%$ under basin ones. The
basin figure is the lower of the two because refreeze is never the majority
class in any basin training fold, so no held-out refreeze lake is ever predicted
correctly.

\textbf{T2, summary statistics}, tests whether temporal modeling is needed at
all. It computes six quantities per channel over days $\le T$: mean, standard
deviation, minimum, maximum, last observed value, and the slope of an ordinary
least squares fit against time, giving $54$ features for nine channels,
classified by the same ridge classifier. It keeps level, spread, extremes and
trend, and discards shape entirely.

\textbf{T3, catch22}~\cite{lubba2019catch22}, tests whether the result depends on
a learned representation. Its $22$ features were selected from a library of
thousands by filtering for classification performance and low mutual redundancy
across the UCR archive, and cover distributional shape, linear and nonlinear
autocorrelation, incremental differences and symbolic dynamics. It is the
standard fixed-feature baseline for this
task~\cite{bagnall2017great,middlehurst2024bakeoff}, applied per channel with the
same classifier.

{\sloppy\textbf{T4, ROCKET}~\cite{dempster2020rocket}, tests whether the result depends
on MiniROCKET's particular restrictions rather than on the kernel family. It is
the transform MiniROCKET simplifies, sampling kernel length, weights, bias,
dilation and padding at random and summarizing each convolution by both its
maximum and its proportion of positive values. We run it at the same kernel
budget.\par}

\textbf{T5, TEASER}~\cite{schafer2020teaser}, tests whether a method built for
earliness finds the same structure. It fits a base classifier at each of a set
of candidate decision points and, at each point, a one-class support vector
machine over the base classifier's class probabilities, committing once that
model accepts the prediction at $v$ consecutive points ($v$ tuned on the training fold). We run it in streaming
mode, removing each instance from the pool as soon as its trigger fires, so
every lake contributes exactly one label and one decision date, under both split
schemes and all five seeds.

\subsection{Evaluation protocol}
\label{sec:protocol}

The protocol is fixed before any result is inspected.

\textbf{Splits.} Two schemes are run at every cutoff. \emph{Random $5$-fold}
cross-validation partitions the $1{,}000$ lakes uniformly at random and is
reported because the systems of Section~\ref{sec:bg-ml} are evaluated this way.
\emph{Leave-one-basin-out} holds out each of the six basins in turn and is our
primary generalization test: lakes within a basin share weather, topography and
drainage infrastructure, so random folds place near-duplicates on both sides of
the split and report an accuracy a model deployed on a new region would not
attain~\cite{roberts2017crossvalidation}. Where the two disagree we report both
and weigh the basin result.

\textbf{Configurations and scale.} The main sweep crosses the $14$ cutoffs with
two preprocessing chains, the leakage-free chain of Section~\ref{sec:causal} and
the conventional one it replaces, and two variable sets, the nine channels of
Table~\ref{tab:channels} alone and those nine plus the five atmospheric channels
of Section~\ref{sec:ablation-carra}. With five seeds and eleven folds this is
$14 \times 2 \times 2 \times 5 \times 11 = 3{,}080$ fits; the baseline arms add
$1{,}078$ more (Appendix~\ref{app:compute}).

\textbf{Metrics.} For a class $c$ with true positives $\mathrm{TP}_c$, false
positives $\mathrm{FP}_c$ and false negatives $\mathrm{FN}_c$,
\begin{equation}
  \mathrm{Rec}_c = \frac{\mathrm{TP}_c}{\mathrm{TP}_c + \mathrm{FN}_c},
  \qquad
  \mathrm{Prec}_c = \frac{\mathrm{TP}_c}{\mathrm{TP}_c + \mathrm{FP}_c},
\end{equation}
\begin{equation}
  F_{1,c} = \frac{2\,\mathrm{Prec}_c\,\mathrm{Rec}_c}
                 {\mathrm{Prec}_c + \mathrm{Rec}_c},
  \qquad
  \mathrm{MacroRec} = \frac{1}{|\mathcal{Y}|}\sum_{c \in \mathcal{Y}}
                      \mathrm{Rec}_c .
\end{equation}
Knowability dates are defined on per-class $F_1$: per class because that is the
quantity being measured, and $F_1$ rather than recall because a class that is
simply over-predicted would reach a recall target without becoming any more
knowable. Aggregate comparisons use macro-recall, which weights the four
outcomes equally and is robust to the class imbalance of the transfer sets in
Section~\ref{sec:transfer}. Accuracy appears only where the class balance makes
it interpretable.

\begin{table*}[!h]
\caption{Per-class $F_1$ (\%) at every prediction cutoff, for the fixed
MiniROCKET and ridge model under the leakage-free preprocessing chain. Each
entry is the mean over five seeds with the standard deviation across seeds. The
bold entry in each row marks the knowability date $T^\ast$, the first cutoff at
which that class reaches the $80\%$ target.}
\label{tab:main}
\footnotesize
\setlength{\tabcolsep}{3pt}
\begin{tabular}{@{}ll*{14}{c}@{}}
\toprule
Split & Class & May 1 & May 15 & Jun 1 & Jun 15 & Jul 1 & Jul 15 & Aug 1 & Aug 15 & Sep 1 & Sep 15 & Oct 1 & Oct 15 & Oct 31 & Dec 31 \\
\midrule
\multirow{4}{*}{Random}
 & Rapid drainage & 54.6\,{\tiny$\pm$0.9} & 57.3\,{\tiny$\pm$0.7} & 57.1\,{\tiny$\pm$1.0} & 61.8\,{\tiny$\pm$1.0} & 72.0\,{\tiny$\pm$1.1} & \textbf{85.0\,{\tiny$\pm$0.8}} & 91.4\,{\tiny$\pm$0.3} & 95.5\,{\tiny$\pm$0.4} & 96.1\,{\tiny$\pm$0.4} & 95.9\,{\tiny$\pm$0.2} & 95.8\,{\tiny$\pm$0.2} & 95.9\,{\tiny$\pm$0.6} & 96.3\,{\tiny$\pm$0.3} & 96.0\,{\tiny$\pm$0.4} \\
 & Slow drainage & 41.6\,{\tiny$\pm$3.6} & 47.0\,{\tiny$\pm$0.8} & 45.8\,{\tiny$\pm$1.5} & 58.2\,{\tiny$\pm$1.7} & 68.4\,{\tiny$\pm$1.0} & 77.1\,{\tiny$\pm$0.9} & \textbf{89.3\,{\tiny$\pm$0.5}} & 92.1\,{\tiny$\pm$0.4} & 93.0\,{\tiny$\pm$0.6} & 93.2\,{\tiny$\pm$0.5} & 93.4\,{\tiny$\pm$0.2} & 93.0\,{\tiny$\pm$0.7} & 93.6\,{\tiny$\pm$0.6} & 93.2\,{\tiny$\pm$0.6} \\
 & Buried & 68.2\,{\tiny$\pm$0.8} & 71.1\,{\tiny$\pm$0.7} & 71.7\,{\tiny$\pm$0.4} & 72.2\,{\tiny$\pm$0.7} & 72.8\,{\tiny$\pm$0.7} & 73.3\,{\tiny$\pm$0.7} & 75.9\,{\tiny$\pm$1.1} & 78.0\,{\tiny$\pm$1.0} & \textbf{81.9\,{\tiny$\pm$0.6}} & 85.6\,{\tiny$\pm$0.9} & 90.0\,{\tiny$\pm$0.4} & 93.8\,{\tiny$\pm$0.6} & 95.2\,{\tiny$\pm$0.8} & 96.8\,{\tiny$\pm$0.5} \\
 & Refreeze & 39.7\,{\tiny$\pm$1.4} & 46.2\,{\tiny$\pm$1.8} & 46.8\,{\tiny$\pm$1.2} & 48.4\,{\tiny$\pm$0.9} & 48.9\,{\tiny$\pm$1.0} & 60.1\,{\tiny$\pm$1.5} & 71.4\,{\tiny$\pm$1.4} & 74.5\,{\tiny$\pm$1.3} & 77.9\,{\tiny$\pm$0.8} & \textbf{82.8\,{\tiny$\pm$1.3}} & 87.5\,{\tiny$\pm$0.6} & 91.1\,{\tiny$\pm$1.2} & 92.2\,{\tiny$\pm$0.5} & 93.6\,{\tiny$\pm$1.0} \\
\midrule
\multirow{4}{*}{Basin}
 & Rapid drainage & 42.7\,{\tiny$\pm$3.8} & 50.8\,{\tiny$\pm$1.4} & 50.0\,{\tiny$\pm$1.4} & 53.2\,{\tiny$\pm$2.4} & 65.1\,{\tiny$\pm$1.5} & \textbf{80.8\,{\tiny$\pm$0.9}} & 89.9\,{\tiny$\pm$0.8} & 95.0\,{\tiny$\pm$0.7} & 95.3\,{\tiny$\pm$0.8} & 95.4\,{\tiny$\pm$0.6} & 95.0\,{\tiny$\pm$0.9} & 94.9\,{\tiny$\pm$0.3} & 96.2\,{\tiny$\pm$0.6} & 95.3\,{\tiny$\pm$0.6} \\
 & Slow drainage & 31.7\,{\tiny$\pm$2.2} & 38.4\,{\tiny$\pm$1.8} & 35.7\,{\tiny$\pm$1.8} & 47.0\,{\tiny$\pm$2.7} & 59.2\,{\tiny$\pm$1.7} & 71.5\,{\tiny$\pm$0.9} & \textbf{88.5\,{\tiny$\pm$0.6}} & 92.0\,{\tiny$\pm$0.4} & 92.4\,{\tiny$\pm$1.8} & 93.3\,{\tiny$\pm$0.6} & 91.2\,{\tiny$\pm$1.1} & 92.0\,{\tiny$\pm$0.6} & 93.0\,{\tiny$\pm$0.3} & 91.2\,{\tiny$\pm$0.5} \\
 & Buried & 66.5\,{\tiny$\pm$1.3} & 69.1\,{\tiny$\pm$1.3} & 68.4\,{\tiny$\pm$0.5} & 68.6\,{\tiny$\pm$1.3} & 70.8\,{\tiny$\pm$1.4} & 70.9\,{\tiny$\pm$0.9} & 73.9\,{\tiny$\pm$1.1} & 74.9\,{\tiny$\pm$1.5} & 79.0\,{\tiny$\pm$0.9} & \textbf{83.7\,{\tiny$\pm$1.1}} & 88.0\,{\tiny$\pm$1.4} & 91.4\,{\tiny$\pm$0.8} & 93.5\,{\tiny$\pm$1.0} & 96.2\,{\tiny$\pm$0.9} \\
 & Refreeze & 33.5\,{\tiny$\pm$0.7} & 41.6\,{\tiny$\pm$1.2} & 41.6\,{\tiny$\pm$0.9} & 42.5\,{\tiny$\pm$1.5} & 43.6\,{\tiny$\pm$1.6} & 53.1\,{\tiny$\pm$1.4} & 69.5\,{\tiny$\pm$0.7} & 70.8\,{\tiny$\pm$1.5} & 75.0\,{\tiny$\pm$1.1} & \textbf{80.7\,{\tiny$\pm$1.3}} & 84.0\,{\tiny$\pm$0.6} & 87.4\,{\tiny$\pm$1.1} & 89.9\,{\tiny$\pm$1.0} & 91.8\,{\tiny$\pm$0.5} \\
\bottomrule
\end{tabular}
\end{table*}

\textbf{Uncertainty.} Confusion matrices are pooled across the folds of a given
seed, and every reported figure is the mean over five seeds with the sample
standard deviation across seeds as the spread, written as $\mu \pm \sigma$. The folds are fixed, so the seed varies only
the random elements of the transform. The catch22 and summary-statistic arms are deterministic given the folds, so no
spread is reported for them. The full-season classifier of
Section~\ref{sec:tstar} is a released model rather than one we refit; for it we
report percentile bootstrap intervals over $1{,}000$ resamples.

\section{Results}
\label{sec:results}

\subsection{Outcomes become knowable in order}
\label{sec:leadtime}

\begin{figure}[!h]
\includegraphics[width=\columnwidth]{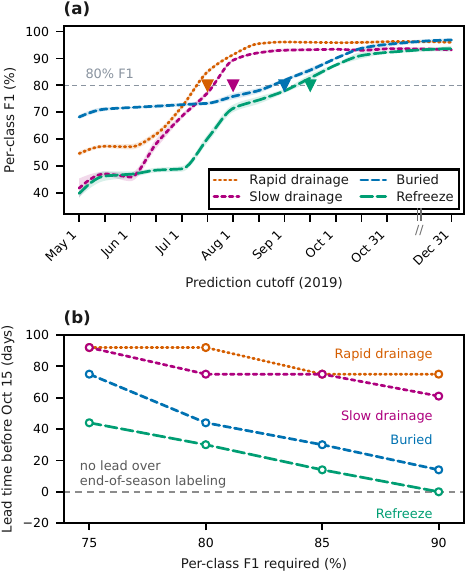}
\caption{(a) Per-class $F_1$ against the prediction cutoff under random folds,
with the $80\%$ target drawn and the knowability date of each class marked on
it. Bands are one standard deviation across the five seeds. (b) Lead time
retained by each class under the same folds as the $F_1$ target $\tau$ is swept from $75\%$ to
$90\%$. Every lead shrinks as the target tightens, and no two classes exchange
places.}
\Description{Two panels. The left panel plots four curves of per-class F1
against calendar date from May to December, with rapid drainage rising first and
refreeze last, and a horizontal line at 80 percent. The right panel plots lead
time in days against the required per-class F1 from 75 to 90 percent, showing
four decreasing lines that do not cross.}
\label{fig:leadtime}
\end{figure}

Table~\ref{tab:main} gives per-class $F_1$ at every cutoff under both split
schemes, and Figure~\ref{fig:leadtime}(a) plots it.\footnote{Code:
\url{https://github.com/ehfahad/supraglacial-leadtime}; data and
results:~\cite{hossain2026artifact}; details in Appendix~\ref{app:repro}.} The four outcomes reach the
$80\%$ target in the order rapid drainage, slow drainage, buried, refreeze.
Under random folds the dates are day $196$ (July~15), day $213$ (August~1), day
$244$ (September~1) and day $258$ (September~15), which are leads of $92$, $75$,
$44$ and $30$ days against the October~15 reference of
Section~\ref{sec:tstar}. Under leave-one-basin-out the two drainage dates are
unchanged and the two storage classes, buried and refreeze, both fall on day
$258$, so the ordering holds but is no longer strict: they become knowable on the same
date once the model must generalize to a new basin.

The ordering follows the physical asymmetry of Section~\ref{sec:bg-lakes}. Rapid
drainage is a sharp, dated event with an unambiguous joint signature, water
fraction collapsing within days while backscatter rises as the exposed bed
roughens, and once that signature is in the record no later observation is
needed to confirm it. Slow drainage is the same transition spread over weeks, so more of the season is
needed to separate a lake that is emptying from one merely shrinking. Buried and
refreeze are defined by an event not having happened, and evidence for a
non-event accumulates only as the season closes without one; refreeze comes last
because freeze-up itself confirms it.

Table~\ref{tab:main} carries two further details. Buried starts high, at
$68.2 \pm 0.8\%$ $F_1$ on May~1 when every other class is below $55\%$, because
buried lakes are distinguished as much by where and what they are as by what
happens to them: they sit high in the percolation zone and carry a distinctive
radar signature from the start of the record, so the slow climb that follows is
the model resolving harder cases rather than discovering the class. And all four
classes are close to flat after day $288$: under random folds macro-recall
gains $1.5$ points between day $288$ and day $365$, so the period after October~15
contributes very little even to a retrospective label.

\begin{table*}[!h]
\caption{Every learner, both split schemes. Left block: macro-recall (\%) at ten
of the fourteen prediction cutoffs under the leakage-free preprocessing chain,
as mean $\pm$ standard deviation over five seeds; catch22 and the summary
statistics are deterministic given the folds and so carry no spread. Right
block: the knowability date $T^\ast$ of each class at $\tau = 0.80$, a calendar
date in 2019, where ``n.r.'' means the target is never reached. Two rows do not
resolve by cutoff. The majority floor ignores the series, and TEASER commits
once per lake at a date of its own choosing, so its accuracy is measured there
and its four dates are medians, not knowability dates; it also runs at $2{,}000$
kernels against $10{,}000$ elsewhere (Appendix~\ref{app:compute}). Bold marks
the best of the remaining four entries in each macro-recall column. All fourteen cutoffs are
in Table~\ref{tab:baselinesfull}.}
\label{tab:allarms}
\footnotesize
\setlength{\tabcolsep}{3pt}
\begin{tabular}{@{}ll*{10}{c}c*{4}{c}@{}}
\toprule
 & & \multicolumn{10}{c}{Macro-recall (\%) at prediction cutoff} & & \multicolumn{4}{c}{Knowability date $T^\ast$} \\
\cmidrule(lr){3-12} \cmidrule(lr){14-17}
Split & Learner & May 1 & Jun 1 & Jul 1 & Jul 15 & Aug 1 & Aug 15 & Sep 1 & Sep 15 & Oct 1 & Dec 31 & & Rapid & Slow & Buried & Refreeze \\
\midrule
\multirow{6}{*}{Random}
 & Majority & \multicolumn{10}{c}{$22.1$, no dependence on the cutoff} & & n.r. & n.r. & n.r. & n.r. \\
 & TEASER & \multicolumn{10}{c}{$59.6 \pm 1.0$ at its own decision dates, median May 15} & & May 15 & May 15 & May 15 & May 15 \\
 & Summary statistics & 51.2 & 54.3 & 63.4 & 72.8 & 78.0 & 79.4 & 79.1 & 82.3 & 82.8 & 81.8 & & Jul 15 & Aug 1 & Sep 1 & Sep 15 \\
 & catch22 & 51.5 & 53.8 & 63.3 & 66.3 & 76.7 & 80.9 & 79.7 & 78.7 & 83.8 & 82.3 & & Aug 1 & Aug 1 & Oct 1 & n.r. \\
 & ROCKET & 45.4\,{\tiny$\pm$0.8} & 52.8\,{\tiny$\pm$0.5} & 64.2\,{\tiny$\pm$0.8} & \textbf{74.8\,{\tiny$\pm$0.8}} & \textbf{82.1\,{\tiny$\pm$0.7}} & \textbf{86.2\,{\tiny$\pm$0.5}} & \textbf{87.3\,{\tiny$\pm$0.6}} & \textbf{90.5\,{\tiny$\pm$0.5}} & \textbf{92.7\,{\tiny$\pm$0.6}} & 94.4\,{\tiny$\pm$0.2} & & Jul 15 & Jul 15 & Sep 1 & Sep 15 \\
 & MiniROCKET & \textbf{52.4\,{\tiny$\pm$1.0}} & \textbf{56.2\,{\tiny$\pm$0.7}} & \textbf{66.1\,{\tiny$\pm$0.8}} & 73.8\,{\tiny$\pm$0.5} & 81.8\,{\tiny$\pm$0.7} & 85.0\,{\tiny$\pm$0.6} & \textbf{87.3\,{\tiny$\pm$0.4}} & 89.4\,{\tiny$\pm$0.5} & 91.7\,{\tiny$\pm$0.2} & \textbf{94.9\,{\tiny$\pm$0.4}} & & Jul 15 & Aug 1 & Sep 1 & Sep 15 \\
\midrule
\multirow{6}{*}{Basin}
 & Majority & \multicolumn{10}{c}{$16.6$, no dependence on the cutoff} & & n.r. & n.r. & n.r. & n.r. \\
 & TEASER & \multicolumn{10}{c}{$50.1 \pm 1.6$ at its own decision dates, median May 15} & & May 15 & Jun 1$^{\dagger}$ & May 15 & May 15 \\
 & Summary statistics & 45.7 & 48.2 & 57.1 & 68.7 & 75.9 & 75.6 & 76.3 & 78.1 & 79.2 & 78.3 & & Aug 1 & n.r. & Sep 1 & Oct 1 \\
 & catch22 & \textbf{46.4} & 45.2 & 56.1 & 61.5 & 70.7 & 76.7 & 75.0 & 75.5 & 78.9 & 75.5 & & Aug 15 & Aug 15 & Oct 31 & n.r. \\
 & ROCKET & 35.9\,{\tiny$\pm$1.2} & 43.2\,{\tiny$\pm$1.2} & 54.8\,{\tiny$\pm$1.7} & 68.2\,{\tiny$\pm$1.5} & 79.1\,{\tiny$\pm$0.8} & \textbf{83.2\,{\tiny$\pm$0.3}} & 85.2\,{\tiny$\pm$0.7} & \textbf{88.6\,{\tiny$\pm$0.5}} & \textbf{91.3\,{\tiny$\pm$0.6}} & \textbf{94.0\,{\tiny$\pm$0.4}} & & Jul 15 & Aug 1 & Sep 15 & Oct 1 \\
 & MiniROCKET & 44.5\,{\tiny$\pm$1.1} & \textbf{49.5\,{\tiny$\pm$0.6}} & \textbf{59.7\,{\tiny$\pm$0.5}} & \textbf{68.8\,{\tiny$\pm$0.7}} & \textbf{80.2\,{\tiny$\pm$0.5}} & 83.1\,{\tiny$\pm$0.9} & \textbf{85.4\,{\tiny$\pm$1.1}} & 88.2\,{\tiny$\pm$0.6} & 89.5\,{\tiny$\pm$0.7} & 93.6\,{\tiny$\pm$0.4} & & Jul 15 & Aug 1 & Sep 15 & Sep 15 \\
\bottomrule
\multicolumn{17}{@{}l}{\footnotesize $^{\dagger}$ TEASER's median decision date
for slow drainage under basin folds is May~15 in three of the five seeds and
June~1 in the other two.}
\end{tabular}
\end{table*}

\subsection{The ordering is model-independent}
\label{sec:modelindep}

{\sloppy If the measured ordering were a consequence of choosing MiniROCKET, it
would not survive a change of learner. The right block of
Table~\ref{tab:allarms} gives the
knowability date of every class under every learner and both split schemes. Four
of the six learners produce a per-cutoff trajectory and therefore a knowability
date, which gives eight learner-by-split cells. The ordering
$T^\ast_{\text{rapid}} \le T^\ast_{\text{slow}} \le T^\ast_{\text{buried}} \le
T^\ast_{\text{refreeze}}$ holds in seven of them, the exception being slow
drainage for the summary-statistic baseline under basin folds, which never
reaches the target at any cutoff. Rapid drainage is earliest in all eight, and
refreeze latest or never reached in all but that one. Under random folds the
summary-statistic baseline recovers all four dates exactly as the fixed model
does, on $54$ features rather than $9{,}996$. The majority-class floor fixes the
other end of the range: it reaches no class target at any cutoff, under either
scheme.\par}

The left block of Table~\ref{tab:allarms} gives the accuracy behind those dates,
and where the learners differ is informative. The two fixed-feature
representations, catch22 and the $54$ summary statistics, track the fixed model
closely through July~1, staying within $1.2$ percentage points of it at May~1
under random folds, then plateau near $83\%$ while the kernel transforms continue
past $94\%$. The early season is therefore carried by a low-dimensional signal,
since level, last value and slope nearly suffice for a drainage event. The
late-season gains come from a different problem, separating buried from
refreeze, which turns on subtler differences in the timing and texture of
freeze-up and does need the richer representation. Hence catch22 never reaches
the refreeze target under either split scheme.

The comparison with ROCKET isolates the kernel restrictions from the kernel
family. The two end the season within $0.5$ percentage points of each other under
both split schemes but separate early: MiniROCKET leads by $7.0$ points at May~1
under random folds and $8.6$ under basin folds, and leads at every cutoff up to
July~1 and August~1 respectively (Table~\ref{tab:baselinesfull}). The cause is
in ROCKET's construction, since with only $121$ days of input its randomly
sampled kernel lengths and dilations
frequently exceed the series length and contribute nothing, whereas MiniROCKET's
dilations follow from the series length and adapt to truncation. Its seed spread
under basin folds is correspondingly wider over the first six cutoffs, $1.2$
points on average against $0.8$, again in Table~\ref{tab:baselinesfull}. Since
the early season is where the measurement lives, we retain MiniROCKET.

The one learner that does not reproduce the ordering is the one built to decide
early, which is the test Section~\ref{sec:bg-ects} promised. TEASER commits on
every lake at a date of its own choosing, and under random folds that date is
May~15 for all four classes, a spread of zero days
against the $62$ that separate the fixed model's earliest and latest knowability
dates under the same folds. Its accuracy at the moment it commits,
$59.6 \pm 1.0\%$ under random folds and $50.1 \pm 1.6\%$ under basin ones, is
close to what the fixed model attains on the same dates, so this is a limit on
expressiveness rather than on quality: a single trigger has one operating point
for four outcomes whose evidence arrives months apart. The cost is uneven across
classes. TEASER misses $50.1 \pm 1.0\%$ of refreeze and
$48.4 \pm 1.5\%$ of slow-drainage lakes against $23.4 \pm 0.7\%$ of buried ones,
and under leave-one-basin-out the slow-drainage error reaches
$65.3 \pm 3.2\%$. A single earliness policy therefore cannot serve all four outcomes
at once, and the per-outcome structure of
Section~\ref{sec:leadtime} cannot be recovered from it.

\subsection{Spatial and temporal transfer}
\label{sec:transfer}

{\sloppy Random cross-validation on spatially structured data is
optimistic~\cite{roberts2017crossvalidation}, so the basin rows of
Tables~\ref{tab:main} and~\ref{tab:allarms} are the ones we ask the
reader to weigh. Under leave-one-basin-out both drainage dates are unchanged, at
July~15 and August~1, and buried moves two weeks later, so the leads of $92$ and
$75$ days do not depend on spatial leakage between folds.\par}

\begin{figure}[!h]
\includegraphics[width=\columnwidth]{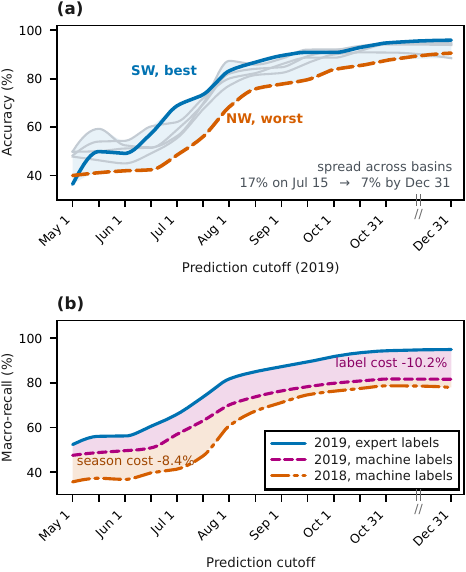}
\caption{(a) Held-out basin accuracy against prediction cutoff. The shaded
envelope spans the six basins; the best and worst basins on July~15 are drawn in
full. (b) Cross-year transfer under random folds, decomposed into the cost of testing on
machine-labeled rather than expert-labeled lakes (Appendix~\ref{app:data}), and the
additional cost of changing melt season. All three settings use the same
model.}
\Description{Two panels. The left panel shows six basin accuracy curves rising
through the season inside a shaded envelope that narrows from a 17-point spread
in mid-July to 7 points by December. The right panel shows three macro-recall
curves for expert-labeled 2019, machine-labeled 2019 and machine-labeled 2018
lakes, with the gaps between them shaded and annotated.}
\label{fig:transfer}
\end{figure}

The cost of spatial transfer is concentrated early. Macro-recall falls from
$52.4 \pm 1.0\%$ to $44.5 \pm 1.1\%$ on May~1 and from $73.8 \pm 0.5\%$ to
$68.8 \pm 0.7\%$ on July~15, and the two schemes converge to within $1.3$ points
by year end. Basins also differ substantially, as Figure~\ref{fig:transfer}(a)
shows: on July~15 held-out basin accuracy spans $17.4$ percentage points, from
$73.5\%$ in Southwest Greenland down to $56.1\%$ in the Northwest, narrowing to
$7.4$ points by December~31. The Northwest is weakest at $12$ of the $14$ cutoffs, consistent with the regional
variability in buried meltwater extent documented across the ice
sheet~\cite{dunmire2021contrasting}.

Transfer to an unseen season entangles two effects: the change of season and
the change from expert to machine labels.
Figure~\ref{fig:transfer}(b) separates them with three settings of the same
model. Testing on the expert-labeled
2019 lakes gives $73.8 \pm 0.5\%$ macro-recall at July~15 and $94.9 \pm 0.4\%$
at December~31. Testing on the machine-labeled lakes of the same season
costs $10.2$ points on average across the $14$ cutoffs, the label
cost; moving to the machine-labeled 2018 lakes costs a further $8.4$, the
season cost, giving $47.4\%$ at July~15 and $78.0\%$ at December~31. The two are
comparable in size, so a cross-year evaluation scored against machine labels
understates transfer by roughly as much as the year change contributes. The
per-class dates do not survive the substitution: under the 2019 machine labels
the drainage classes swap and refreeze never reaches the target, and on the 2018
set only buried reaches it, on October~31. Knowability is measurable from the
expert labels, not from the machine ones.

\subsection{Error modes explain the ordering}
\label{sec:errormodes}

\begin{figure}[!h]
\includegraphics[width=\columnwidth]{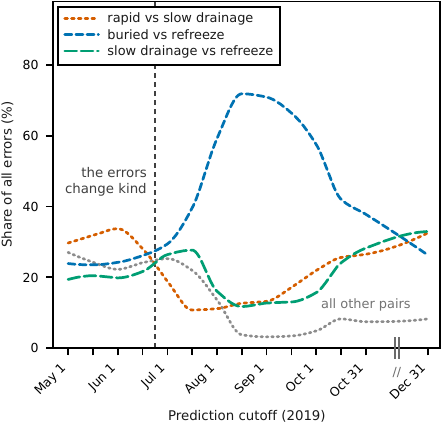}
\caption{Share of total misclassifications carried by each of the three dominant
class pairs, against the prediction cutoff under random folds, with the remaining pairs pooled as
the dotted line. The vertical line marks where the leading pair changes: the
drainage pair dominates before it and resolves first, the storage pair after,
and the storage pair still carries two thirds of the error on September~15.}
\Description{Three curves showing the fraction of total errors attributable to
each of three class pairs over the melt season. The rapid versus slow drainage
pair dominates early and falls away, while the buried versus refreeze pair
persists into September.}
\label{fig:errormodes}
\end{figure}

Figure~\ref{fig:errormodes} resolves the confusion mass at each cutoff into the
three class pairs that carry most of it, and the result matches the
physical account of Section~\ref{sec:leadtime}. Early in the season the dominant
confusion is rapid against slow drainage: both are lakes that have begun to lose
water, and only the rate distinguishes them, which takes several weeks of record
to estimate. That pair resolves first, and its resolution carries rapid drainage
over the target on July~15.

The confusion that persists is buried against refreeze, with slow drainage
against refreeze beneath it. Both follow from one cause: until the season closes, a lake that will refreeze,
one that will be buried and one draining very slowly all simply persist, and
persistence is what the record shows. The buried and refreeze pair still contributes error on September~15, which is why the
storage classes reach the target weeks after rapid drainage does.
Drainage therefore separates from non-drainage well before one kind of
non-drainage separates from the other.

\section{Ablations}
\label{sec:ablations}

\subsection{Cost of removing leakage}
\label{sec:ablation-leak}

Running the identical sweep under both chains
(Sections~\ref{sec:bg-leak} and~\ref{sec:causal}) measures what the trailing
requirement costs and what the conventional chain gains from reading ahead
(Table~\ref{tab:leak}).

\begin{table}[!h]
\caption{Macro-recall (\%) under the leakage-free chain and the conventional
chain that reads across gaps and window edges, mean $\pm$ standard deviation
over five seeds. The better of the two chains in each column is bold; the
largest advantage the conventional chain obtains at any cutoff is $1.3$
points.}
\label{tab:leak}
\small
\setlength{\tabcolsep}{3pt}
\begin{tabular}{@{}llrrrr@{}}
\toprule
Split & Chain & May 1 & Jul 1 & Sep 1 & Dec 31 \\
\midrule
\multirow{2}{*}{Random}
 & Leakage-free  & \textbf{52.4 $\pm$ 1.0} & 66.1 $\pm$ 0.8 & 87.3 $\pm$ 0.4 & \textbf{94.9 $\pm$ 0.4} \\
 & Conventional  & 51.1 $\pm$ 1.1 & \textbf{66.4 $\pm$ 0.9} & \textbf{87.9 $\pm$ 0.4} & 94.4 $\pm$ 0.6 \\
\midrule
\multirow{2}{*}{Basin}
 & Leakage-free  & 44.5 $\pm$ 1.1 & \textbf{59.7 $\pm$ 0.5} & 85.4 $\pm$ 1.1 & \textbf{93.6 $\pm$ 0.4} \\
 & Conventional  & \textbf{45.4 $\pm$ 1.1} & 59.0 $\pm$ 0.4 & \textbf{86.3 $\pm$ 0.8} & 93.4 $\pm$ 0.3 \\
\bottomrule
\end{tabular}
\end{table}

The two chains are close everywhere. Under random folds the leakage-free chain
is ahead by $3.4$ points in mid-May and behind by at most $1.2$ in
mid-September, a mean absolute difference of $0.9$ points across the $14$
cutoffs; under basin folds it is ahead at nine of the $14$, by as much as $2.6$
points in mid-May, and behind by at most $1.3$. No knowability date changes
under either chain (Table~\ref{tab:mainfull}).

Two conclusions follow. The measurement of Section~\ref{sec:leadtime} carries
no dependence on future observations and gives up little accuracy for that guarantee;
and what the conventional chain gains from reading across gaps and window edges
is worth at most $1.3$ percentage points, so published retrospective accuracies
are not materially inflated by it. The dependence is a hazard for the temporal
question rather than a defect in the retrospective results, but any pipeline
that may one day be
asked when something became knowable should use trailing operators from the
start.

\subsection{Atmospheric reanalysis channels}
\label{sec:ablation-carra}

Prior work on this record pairs the satellite series with atmospheric
reanalysis: Hossain et al.~\cite{hossain2025rictsc} draw daily fields from the
Copernicus Arctic Regional Reanalysis (CARRA), a $2.5$\,km reanalysis of the
European Arctic~\cite{schyberg2020carra}, at the grid cell nearest each lake.
Atmospheric forcing is also the natural candidate for an early predictive
signal, since melt production and snowfall drive a lake toward one outcome or
another, so we test it rather than assume it. We add five CARRA fields
at the cadence of the satellite channels, surface meltwater runoff
(\chan{runoff}), broadband albedo (\chan{albedo}), relative humidity at $2$\,m
(\chan{rh2m}), snow water equivalent (\chan{sde\textsubscript{swe}}) and surface
pressure (\chan{sp}), holding everything else fixed.

Adding them does not help. Under random folds macro-recall moves from $+0.3$
points in mid-May to $-1.6$ on July~1, a mean of $-0.7$, and no knowability date
changes; under basin folds the mean is $-1.1$, the worst single loss $2.3$
points, and two dates move later, rapid drainage to August~1 and refreeze to
October~1. They are therefore not in the pipeline of
Section~\ref{sec:method}: they add five channels and, where they change
anything, cost accuracy.

The measurement explains part of the null. A two-way variance decomposition
(Appendix~\ref{app:results}) shows two of the five channels to be almost static
across the season within a lake, snow water equivalent carrying $92.3\%$ of its
variance between lakes and surface pressure $97.5\%$. Both act as near-constant
lake descriptors rather than temporal signal. That does not extend to the other three,
whose between-lake fractions of $13.4$ to $22.6\%$ sit inside the satellite
range, and a linear reconstruction of the five from the nine reaches only
$R^2 = 0.02$ to $0.52$.
The remainder is plausibly scale: most lakes in the reference set are smaller
than one CARRA grid cell, so its channels describe the region rather than the
lake.

\subsection{Sensitivity to the target}
\label{sec:ablation-tau}

The $80\%$ target of Section~\ref{sec:tstar} is a choice, so we sweep it.
Figure~\ref{fig:leadtime}(b) plots the lead each class retains as $\tau$ rises
from $0.75$ to $0.90$. Every lead shrinks monotonically, as it must, and the
four curves never cross, so the ordering survives at every target tested: at
$\tau = 0.75$ both drainage classes retain $92$ days and refreeze $44$, while at
$\tau = 0.90$ rapid drainage retains $75$ days and refreeze retains none at all.
Under random folds at the $80\%$ target all five seeds give the same dates; at the
other targets, and under basin folds, single seeds shift some dates by one or two
cutoffs.

\section{Limitations}
\label{sec:limitations}

\textbf{One ice sheet, two melt seasons.} All results come from Greenland in 2018
and 2019, and 2018 enters only as a transfer target, so the dates are specific to
these seasons. A warmer or cooler season would shift them by an amount two years
cannot establish.

\textbf{Label provenance outside the reference set.} The measurement rests on
$1{,}000$ expert-labeled lakes; the transfer experiments extend to $8{,}992$
further lakes labeled by the pipeline of Dunmire et
al.~\cite{dunmire2024greenland}, and scoring on those lakes costs $10.2$
points of macro-recall even within the same season. Section~\ref{sec:transfer} separates that cost from
model error but cannot remove it, so the cross-year conclusion is the weaker
one.

\textbf{The reference set is balanced by construction.} With $250$ lakes per
class, the accuracies here are not deployment accuracies on the true, unbalanced
population, and because per-class $F_1$ depends on precision, the dates would
also shift under a realistic class mix.

\textbf{Knowability is a property of a decision rule.} We define it as the date
at which a classifier of a specified family reaches a specified accuracy, so a
stronger model or a different target moves it.
Sections~\ref{sec:modelindep} and~\ref{sec:ablation-tau} show the ordering
survives both; the dates are not constants.

\textbf{Two-week resolution.} With $14$ cutoffs a date is resolved to roughly
two weeks. A daily grid would sharpen the dates at roughly $18$ times the compute, and
the ordering the argument rests on is already established at this resolution.

\section{Conclusion}
\label{sec:conclusion}

Supraglacial lake outcomes are labeled retrospectively because of how the
systems producing them are built, not because the observations require it. This
paper separates the two by truncating the input and retraining at each of $14$
cutoffs, and finds that the four become knowable in a fixed order whose shape
follows the physics: a dated event first, the same event drawn out over weeks
next, and the two outcomes defined by an event failing to occur last. Two of the
four resolve months before the season closes, rapid drainage by July~15 and slow
drainage by August~1, respectively $92$ and $75$ days ahead of the earliest date a full-season
pipeline can be computed at all. The ordering holds in seven of eight
learner-by-split cells, across learners spanning an $18$-point end-of-season accuracy range, under
leave-one-basin-out evaluation and at every target we tested, and it is measured
on features that never read the future, for at most $1.3$ percentage points. It
does not survive being scored against machine labels in a second
season, where label quality and the lakes themselves both change.

For an operational system this means there should not be one release date. Rapid
drainage, the outcome that matters most for ice dynamics, reaches the target on
July~15, three months before a full-season pipeline can be computed at all,
whereas refreeze does not reach it before mid-September. A single release date
either discards roughly three months of lead on the outcome that most needs it
or claims a confidence in the storage outcomes that the record does not yet
carry.

Two directions follow. More expert-labeled seasons would show whether these
dates are stable, drift with a warming climate, or vary by basin, and the
$17.4$-point spread across basins makes the last worth testing first. And since
most existing early classification methods commit under one policy for all classes, a
per-outcome trigger function is the natural instrument; the curves reported here
are its calibration target.

\begin{acks}
We thank Devon Dunmire (University at Buffalo), Hammad Younas (St.~John's
School) and Brendan Myers (National Center for Atmospheric Research) for
collecting, labeling and preparing the supraglacial lake dataset used in this
work.

This work is supported by iHARP: NSF HDR Institute for Harnessing Data and
Model Revolution in the Polar Regions (Award\# 2118285). The views expressed in
this work do not necessarily reflect the policies of the NSF, and endorsement by
the Federal Government should not be inferred.
\end{acks}

\balance
\bibliographystyle{ACM-Reference-Format}
\bibliography{references}

\clearpage
\appendix
\section*{APPENDIX}

\section{Reproducibility}
\label{app:repro}

\textbf{Inputs.} The satellite record, the expert labels and the released
full-season models are those of Dunmire et al.~\cite{dunmire2024greenland},
deposited under CC-BY 4.0~\cite{dunmire2024data}. The CARRA reanalysis is available
from the Copernicus Climate Data Store~\cite{schyberg2020carra}. Neither is
re-hosted here; both are fetched by DOI.

{\sloppy\textbf{Code.} Code for the full pipeline, from the inputs above to
the reported results, is released under the MIT license at
\url{https://github.com/ehfahad/supraglacial-leadtime}.\par}

{\sloppy\textbf{Data and results.} The preprocessed record and the complete result
record are deposited separately under CC-BY 4.0~\cite{hossain2026artifact}.
\texttt{processed.zip} holds the leakage-free and conventional tensors for both
seasons, each carrying the fold and basin assignment, the expert and machine
labels and the drainage date of every lake, so the exact splits and seeds of
Section~\ref{sec:protocol} need not be regenerated. \texttt{results.zip} holds
the per-fold, per-seed record of all $3{,}080$ fits of the main sweep and of
each baseline arm, the per-lake decision dates of all $10{,}000$ TEASER
decisions, the per-fold majority-class floor and the cross-year evaluation.
Every number in every table of this paper is recomputable from the deposit
alone. Reproducing the main sweep from the processed record requires no
specialized hardware and completes in the budget of Appendix~\ref{app:compute}.\par}

The deposited tensors contain modified Copernicus Climate Change Service
information (2026); neither the European Commission nor ECMWF is responsible for
any use of it.

\section{Dataset Details}
\label{app:data}

\textbf{Class composition.} The expert-labeled reference set contains exactly
$250$ lakes in each class, but its per-basin composition is uneven, which is why
leave-one-basin-out folds differ in size and in difficulty: Southwest Greenland
contributes $245$ lakes of which $89$ are rapid drainage, whereas Southeast
Greenland contributes $92$ of which $10$ are. Table~\ref{tab:basins} gives the
full breakdown.

\begin{table}[!h]
\caption{Composition of the six leave-one-basin-out test folds over the
$1{,}000$ expert-labeled lakes.}
\label{tab:basins}
\small
\begin{tabular}{@{}lrrrrr@{}}
\toprule
Basin & Lakes & Rapid & Slow & Buried & Refreeze \\
\midrule
Central West & 208 & 60 & 47 & 44 & 57 \\
Northeast    & 161 & 27 & 43 & 53 & 38 \\
North        & 141 & 21 & 39 & 41 & 40 \\
Northwest    & 153 & 43 & 22 & 40 & 48 \\
Southeast    &  92 & 10 & 31 & 30 & 21 \\
Southwest    & 245 & 89 & 68 & 42 & 46 \\
\midrule
Total        & 1000 & 250 & 250 & 250 & 250 \\
\bottomrule
\end{tabular}
\end{table}

\textbf{Machine-labeled lakes.} The record also carries labels produced by the
pipeline of Dunmire et al.~\cite{dunmire2024greenland} for every detected lake
in both seasons. These are never used for training or for any headline result;
they are the transfer targets of Section~\ref{sec:transfer}, and
Table~\ref{tab:machine} gives their composition. Two properties matter for
reading that section. The class balance is very different from the expert set:
slow drainage is the plurality class in 2019 at $42.7\%$ of lakes, whereas
refreeze is the plurality class in 2018 at $34.6\%$, consistent with 2018 being
the cooler season~\cite{dunmire2024greenland,subramanian2024fate}. And the basin mix differs between years, with Northeast
Greenland contributing $26.8\%$ of machine-labeled lakes in 2019 against
$18.3\%$ in 2018. Accuracy on these sets therefore reflects a different
population as well as different label quality.

\begin{table*}[!h]
\caption{Per-class $F_1$ (\%) at every cutoff under the \emph{conventional}
preprocessing chain, the counterpart to Table~\ref{tab:main}. Mean over five
seeds with the standard deviation across seeds; bold marks the knowability date.
No class's knowability date differs from the leakage-free chain, which is the
detail behind the summary in Section~\ref{sec:ablation-leak}.}
\label{tab:mainfull}
\footnotesize
\setlength{\tabcolsep}{3pt}
\begin{tabular}{@{}ll*{14}{c}@{}}
\toprule
Split & Class & May 1 & May 15 & Jun 1 & Jun 15 & Jul 1 & Jul 15 & Aug 1 & Aug 15 & Sep 1 & Sep 15 & Oct 1 & Oct 15 & Oct 31 & Dec 31 \\
\midrule
\multirow{4}{*}{Random}
 & Rapid drainage & 54.1\,{\tiny$\pm$0.8} & 54.7\,{\tiny$\pm$1.6} & 57.0\,{\tiny$\pm$1.0} & 62.1\,{\tiny$\pm$1.1} & 75.1\,{\tiny$\pm$1.1} & \textbf{85.4\,{\tiny$\pm$0.8}} & 91.4\,{\tiny$\pm$0.8} & 95.5\,{\tiny$\pm$0.2} & 95.8\,{\tiny$\pm$0.2} & 95.6\,{\tiny$\pm$0.4} & 95.5\,{\tiny$\pm$0.4} & 95.6\,{\tiny$\pm$0.4} & 95.4\,{\tiny$\pm$0.8} & 95.8\,{\tiny$\pm$0.5} \\
 & Slow drainage & 41.5\,{\tiny$\pm$1.7} & 43.3\,{\tiny$\pm$1.6} & 46.4\,{\tiny$\pm$1.2} & 57.5\,{\tiny$\pm$2.0} & 68.6\,{\tiny$\pm$1.3} & 78.3\,{\tiny$\pm$1.0} & \textbf{90.1\,{\tiny$\pm$1.0}} & 91.4\,{\tiny$\pm$0.7} & 92.8\,{\tiny$\pm$0.7} & 92.5\,{\tiny$\pm$0.2} & 92.0\,{\tiny$\pm$0.8} & 92.5\,{\tiny$\pm$0.6} & 91.9\,{\tiny$\pm$1.0} & 92.3\,{\tiny$\pm$1.1} \\
 & Buried & 65.3\,{\tiny$\pm$1.2} & 68.3\,{\tiny$\pm$1.0} & 71.2\,{\tiny$\pm$1.4} & 71.3\,{\tiny$\pm$0.8} & 73.4\,{\tiny$\pm$0.6} & 74.6\,{\tiny$\pm$0.7} & 75.4\,{\tiny$\pm$1.5} & 79.2\,{\tiny$\pm$1.0} & \textbf{83.3\,{\tiny$\pm$0.7}} & 88.5\,{\tiny$\pm$0.7} & 92.1\,{\tiny$\pm$0.6} & 94.2\,{\tiny$\pm$0.6} & 95.4\,{\tiny$\pm$0.4} & 96.4\,{\tiny$\pm$0.5} \\
 & Refreeze & 38.5\,{\tiny$\pm$2.3} & 41.2\,{\tiny$\pm$3.2} & 44.9\,{\tiny$\pm$1.0} & 44.6\,{\tiny$\pm$1.4} & 44.9\,{\tiny$\pm$1.2} & 61.4\,{\tiny$\pm$1.2} & 71.8\,{\tiny$\pm$1.2} & 74.9\,{\tiny$\pm$1.9} & 79.8\,{\tiny$\pm$0.4} & \textbf{85.7\,{\tiny$\pm$1.1}} & 89.6\,{\tiny$\pm$0.6} & 92.1\,{\tiny$\pm$0.4} & 92.3\,{\tiny$\pm$0.6} & 93.0\,{\tiny$\pm$0.7} \\
\midrule
\multirow{4}{*}{Basin}
 & Rapid drainage & 46.9\,{\tiny$\pm$1.7} & 48.5\,{\tiny$\pm$1.0} & 48.0\,{\tiny$\pm$1.5} & 48.9\,{\tiny$\pm$0.6} & 66.7\,{\tiny$\pm$1.1} & \textbf{82.4\,{\tiny$\pm$1.4}} & 89.2\,{\tiny$\pm$0.9} & 94.7\,{\tiny$\pm$1.0} & 96.2\,{\tiny$\pm$0.3} & 96.0\,{\tiny$\pm$0.4} & 95.2\,{\tiny$\pm$0.7} & 95.7\,{\tiny$\pm$0.4} & 95.5\,{\tiny$\pm$0.6} & 95.3\,{\tiny$\pm$0.9} \\
 & Slow drainage & 33.8\,{\tiny$\pm$2.0} & 36.9\,{\tiny$\pm$1.4} & 37.6\,{\tiny$\pm$0.1} & 48.8\,{\tiny$\pm$1.1} & 60.1\,{\tiny$\pm$1.5} & 72.3\,{\tiny$\pm$2.0} & \textbf{87.6\,{\tiny$\pm$1.1}} & 91.5\,{\tiny$\pm$1.0} & 93.3\,{\tiny$\pm$1.0} & 93.0\,{\tiny$\pm$0.5} & 91.7\,{\tiny$\pm$1.0} & 92.4\,{\tiny$\pm$0.8} & 91.8\,{\tiny$\pm$0.6} & 91.1\,{\tiny$\pm$0.7} \\
 & Buried & 60.3\,{\tiny$\pm$0.9} & 64.0\,{\tiny$\pm$1.3} & 65.8\,{\tiny$\pm$1.3} & 66.9\,{\tiny$\pm$1.2} & 68.3\,{\tiny$\pm$0.5} & 68.2\,{\tiny$\pm$1.6} & 71.1\,{\tiny$\pm$1.9} & 74.3\,{\tiny$\pm$1.9} & 79.6\,{\tiny$\pm$1.5} & \textbf{86.1\,{\tiny$\pm$1.5}} & 89.2\,{\tiny$\pm$1.9} & 91.8\,{\tiny$\pm$0.9} & 94.3\,{\tiny$\pm$1.3} & 95.8\,{\tiny$\pm$0.8} \\
 & Refreeze & 37.0\,{\tiny$\pm$2.2} & 39.0\,{\tiny$\pm$1.6} & 40.0\,{\tiny$\pm$1.5} & 38.3\,{\tiny$\pm$2.0} & 40.3\,{\tiny$\pm$1.5} & 53.8\,{\tiny$\pm$1.6} & 67.1\,{\tiny$\pm$1.5} & 71.1\,{\tiny$\pm$1.0} & 76.6\,{\tiny$\pm$1.0} & \textbf{83.0\,{\tiny$\pm$1.4}} & 85.7\,{\tiny$\pm$1.6} & 88.3\,{\tiny$\pm$1.2} & 90.4\,{\tiny$\pm$1.6} & 91.5\,{\tiny$\pm$1.1} \\
\bottomrule
\end{tabular}
\end{table*}

\begin{table*}[!h]
\caption{Macro-recall (\%) at \emph{every} prediction cutoff for the four
learners that produce a per-cutoff curve, under the leakage-free preprocessing
chain, given as mean $\pm$ standard deviation over five seeds. This is the full
version of the left block of Table~\ref{tab:allarms}, which shows ten of the
fourteen cutoffs. The majority-class floor and TEASER are omitted here because
neither varies with the cutoff: the first ignores the series, the second commits
once per lake at a date of its own choosing. The catch22 and summary-statistic
pipelines are deterministic given the folds, so no spread is reported for them.
The best entry in each column is bold. The trivial floor is $25.0\%$
macro-recall.}
\label{tab:baselinesfull}
\footnotesize
\setlength{\tabcolsep}{3pt}
\begin{tabular}{@{}ll*{14}{c}@{}}
\toprule
Split & Learner & May 1 & May 15 & Jun 1 & Jun 15 & Jul 1 & Jul 15 & Aug 1 & Aug 15 & Sep 1 & Sep 15 & Oct 1 & Oct 15 & Oct 31 & Dec 31 \\
\midrule
\multirow{4}{*}{Random}
 & Summary statistics & 51.2 & 53.0 & 54.3 & 57.5 & 63.4 & 72.8 & 78.0 & 79.4 & 79.1 & 82.3 & 82.8 & 83.0 & 83.0 & 81.8 \\
 & catch22 & 51.5 & 54.8 & 53.8 & 54.3 & 63.3 & 66.3 & 76.7 & 80.9 & 79.7 & 78.7 & 83.8 & 84.5 & 83.7 & 82.3 \\
 & ROCKET & 45.4\,{\tiny$\pm$0.8} & 51.3\,{\tiny$\pm$0.5} & 52.8\,{\tiny$\pm$0.5} & 57.2\,{\tiny$\pm$0.4} & 64.2\,{\tiny$\pm$0.8} & \textbf{74.8\,{\tiny$\pm$0.8}} & \textbf{82.1\,{\tiny$\pm$0.7}} & \textbf{86.2\,{\tiny$\pm$0.5}} & \textbf{87.3\,{\tiny$\pm$0.6}} & \textbf{90.5\,{\tiny$\pm$0.5}} & \textbf{92.7\,{\tiny$\pm$0.6}} & \textbf{93.6\,{\tiny$\pm$0.5}} & 94.2\,{\tiny$\pm$0.1} & 94.4\,{\tiny$\pm$0.2} \\
 & MiniROCKET & \textbf{52.4\,{\tiny$\pm$1.0}} & \textbf{56.2\,{\tiny$\pm$0.6}} & \textbf{56.2\,{\tiny$\pm$0.7}} & \textbf{60.7\,{\tiny$\pm$0.6}} & \textbf{66.1\,{\tiny$\pm$0.8}} & 73.8\,{\tiny$\pm$0.5} & 81.8\,{\tiny$\pm$0.7} & 85.0\,{\tiny$\pm$0.6} & 87.3\,{\tiny$\pm$0.4} & 89.4\,{\tiny$\pm$0.5} & 91.7\,{\tiny$\pm$0.2} & 93.4\,{\tiny$\pm$0.6} & \textbf{94.3\,{\tiny$\pm$0.2}} & \textbf{94.9\,{\tiny$\pm$0.4}} \\
\midrule
\multirow{4}{*}{Basin}
 & Summary statistics & 45.7 & 47.3 & 48.2 & 51.8 & 57.1 & 68.7 & 75.9 & 75.6 & 76.3 & 78.1 & 79.2 & 80.5 & 78.8 & 78.3 \\
 & catch22 & \textbf{46.4} & 49.1 & 45.2 & 51.2 & 56.1 & 61.5 & 70.7 & 76.7 & 75.0 & 75.5 & 78.9 & 77.4 & 77.9 & 75.5 \\
 & ROCKET & 35.9\,{\tiny$\pm$1.2} & 38.8\,{\tiny$\pm$0.8} & 43.2\,{\tiny$\pm$1.2} & 49.2\,{\tiny$\pm$0.6} & 54.8\,{\tiny$\pm$1.7} & 68.2\,{\tiny$\pm$1.5} & 79.1\,{\tiny$\pm$0.8} & \textbf{83.2\,{\tiny$\pm$0.3}} & 85.2\,{\tiny$\pm$0.7} & \textbf{88.6\,{\tiny$\pm$0.5}} & \textbf{91.3\,{\tiny$\pm$0.6}} & \textbf{93.1\,{\tiny$\pm$0.8}} & \textbf{93.5\,{\tiny$\pm$0.2}} & \textbf{94.0\,{\tiny$\pm$0.4}} \\
 & MiniROCKET & 44.5\,{\tiny$\pm$1.1} & \textbf{50.4\,{\tiny$\pm$0.4}} & \textbf{49.5\,{\tiny$\pm$0.6}} & \textbf{53.1\,{\tiny$\pm$1.7}} & \textbf{59.7\,{\tiny$\pm$0.5}} & \textbf{68.8\,{\tiny$\pm$0.7}} & \textbf{80.2\,{\tiny$\pm$0.5}} & 83.1\,{\tiny$\pm$0.9} & \textbf{85.4\,{\tiny$\pm$1.1}} & 88.2\,{\tiny$\pm$0.6} & 89.5\,{\tiny$\pm$0.7} & 91.4\,{\tiny$\pm$0.6} & 93.1\,{\tiny$\pm$0.4} & 93.6\,{\tiny$\pm$0.4} \\
\bottomrule
\end{tabular}
\end{table*}

\begin{table}[!h]
\caption{Machine-labeled lakes by basin and season, excluding the $1{,}000$
expert-labeled lakes of Table~\ref{tab:basins}. No expert labels exist for 2018,
so that season is machine-labeled throughout. Both sets are transfer targets
only.}
\label{tab:machine}
\small
\setlength{\tabcolsep}{4pt}
\begin{tabular}{@{}llrrrrr@{}}
\toprule
Season & Basin & Lakes & Rapid & Slow & Buried & Refreeze \\
\midrule
\multirow{7}{*}{2019}
 & Central West &  792 &  225 &  341 &  107 &  119 \\
 & Northeast    & 1379 &  266 &  451 &  369 &  293 \\
 & North        &  613 &   92 &  278 &  124 &  119 \\
 & Northwest    &  884 &  160 &  368 &  208 &  148 \\
 & Southeast    &  199 &   44 &  105 &   25 &   25 \\
 & Southwest    & 1279 &  389 &  652 &   98 &  140 \\
 & Total        & 5146 & 1176 & 2195 &  931 &  844 \\
\midrule
\multirow{7}{*}{2018}
 & Central West &  679 &  118 &  236 &  172 &  153 \\
 & Northeast    &  705 &  125 &  114 &  126 &  340 \\
 & North        &  539 &  121 &   86 &   61 &  271 \\
 & Northwest    &  633 &   64 &  135 &  186 &  248 \\
 & Southeast    &  213 &   35 &   52 &   55 &   71 \\
 & Southwest    & 1077 &  247 &  393 &  190 &  247 \\
 & Total        & 3846 &  710 & 1016 &  790 & 1330 \\
\bottomrule
\end{tabular}
\end{table}

\textbf{Leading-value constants.} Last observation carried forward cannot fill
the interval before a lake's first observation. Those days take fixed constants
that depend on no data: $-25$\,dB for \chan{HV\textsubscript{lake}} and
\chan{HV\textsubscript{out}}, typical dry-snow backscatter; $0$ for
\chan{HV\textsubscript{anom}} and for all water fraction and probability
channels; $260$\,K for \chan{t2m}; and $90^\circ$ for both zenith channels,
which encodes an unobservable surface. In the ablation of
Section~\ref{sec:ablation-carra} the five atmospheric channels take $0$ for
\chan{runoff} and \chan{sde\textsubscript{swe}}, $0.8$ for \chan{albedo}, $80\%$
for \chan{rh2m} and $100$\,kPa for \chan{sp}. The same constants are used under
both preprocessing chains.

\textbf{Cutoff grid.} The grid places every cutoff on the 1st or 15th of a month
from May~1 to October~15 and adds October~31, the end of the melt season, and
December~31, a full-year reference. Day $121$ is a hard floor because the
processed record begins there. The earliest drainage in the expert set falls on
day $129$ and the first percentile of event dates on day $142$, so a grid
starting later than May~1 would discard the period in which the earliest events
occur.

\section{Implementation and Compute}
\label{app:compute}

\textbf{Software.} MiniROCKET, ROCKET, catch22 and TEASER are used as
implemented in sktime~\cite{loning2019sktime}; the ridge classifier, its exact
leave-one-out regularization search and the scoring metrics are from
scikit-learn~\cite{pedregosa2011scikit}.

\textbf{Fit counts.} Beyond the $3{,}080$ fits of the main sweep, the ROCKET arm
adds $14 \times 5 \times 11 = 770$ fits and the catch22 and summary-statistic
arms $154$ each, both being deterministic given the folds. TEASER is run for
every fold of both split schemes at all five seeds, $5 \times 11 = 55$ runs
yielding $10{,}000$ decisions, one per lake per seed and scheme. The
majority-class floor fits nothing: it takes the modal training label and its cost
is not measurable at this resolution.

\textbf{Hyperparameters.} Kernel count is $10{,}000$ for MiniROCKET and ROCKET.
TEASER uses $2{,}000$ kernels in its internal transform, because it fits both a
base classifier and a one-class classifier with a parameter grid at each of the
$14$ decision points and the full kernel count would multiply its cost by
roughly five without changing the comparison; this reduction is the one
deviation from matched settings across arms and is recorded here for that
reason. The ridge regularization grid is ten logarithmically spaced values in
$[10^{-3}, 10^{3}]$, selected within each training fold by the closed-form
leave-one-out identity. Smoothing windows are $12$ days for all channels. The
Hampel filter uses a trailing window of $12$ days and a threshold of $3$ scaled
median absolute deviations. Seeds are $42$ through $46$ throughout. No
hyperparameter was selected using test-fold data at any point.

\textbf{Hardware and cost.} All runs were executed as SLURM array jobs on the
\texttt{general} partition of the UMBC CHIP CPU cluster, whose nodes carry $36$
or $48$ physical cores. Each array task requested $16$ cores and $48$ to
$64$\,GB of memory; no GPU was used at any stage. Measured solver time, summed
over every fit and excluding data loading, is $4.8$ core-hours for the $3{,}080$
fits of the main sweep, $2.1$ for the ROCKET arm, $5.0$ for the two catch22
arms, $9.8$ for TEASER and $8.3$ for the cross-year evaluation, approximately
$30$ core-hours in total. Wall-clock time for the main sweep is about $70$
minutes on $16$ workers. The dominant cost is TEASER, at roughly $650$ seconds
per run and $9.8$ core-hours over its $55$ runs, because it refits at every decision
point.

\section{Additional Results}
\label{app:results}

Table~\ref{tab:mainfull} repeats Table~\ref{tab:main} under the conventional
preprocessing chain, so the two can be compared cutoff by cutoff rather than at
the four dates summarized in Section~\ref{sec:ablation-leak}. The released
record additionally contains, for every cutoff, split, seed and fold, the full
$4 \times 4$ confusion matrix and the per-class precision, recall and $F_1$
derived from it, for the main sweep under both preprocessing chains and both
variable sets, and for each baseline arm. Three items referenced from the body
are summarized here.

\textbf{Per-basin behavior.} Held-out accuracy on July~15 spans $17.4$
percentage points across the six basins and narrows to $7.4$ points by
December~31. Northwest Greenland is the weakest basin at $12$ of the $14$
cutoffs and Southwest Greenland the strongest at $6$.

\textbf{The variance decomposition} behind Section~\ref{sec:ablation-carra}
splits each channel's variance into between-lake, between-day and residual
components across the $1{,}000$ reference lakes and the $168$ days from May~1 to
October~15. The between-lake fraction is $92.3\%$ for
\chan{sde\textsubscript{swe}} and $97.5\%$ for \chan{sp}, against $13.4\%$ for
\chan{runoff}, $19.9\%$ for \chan{rh2m} and $22.6\%$ for \chan{albedo}; the nine
input channels of Table~\ref{tab:channels} span $11.6$ to $32.7\%$. Linear
reconstruction of each CARRA channel from those nine attains $R^2$ between
$0.02$ and $0.52$, highest for \chan{albedo} and lowest for
\chan{sde\textsubscript{swe}}.

\textbf{The full-season reference pipeline} of Section~\ref{sec:tstar} is
evaluated under truncation with percentile bootstrap confidence intervals over
$1{,}000$ resamples. Its Sentinel-1 component attains $21.0\%$ accuracy
($95\%$ CI $15.5$ to $26.5$) on May~1 against a $25.0\%$ chance level on its
four-class label set, and its Sentinel-2 component $32.9\%$ ($95\%$ CI $25.9$ to
$39.9$) against a $33.3\%$ chance level on its three-class label set. Neither
component is distinguishable from chance before mid-June: both bootstrap
intervals contain the chance level at every cutoff up to June~1.

\end{document}